\documentclass[10pt, a4paper]{article}

\usepackage{lrec2026}

\usepackage{listings}
\usepackage{listingsutf8} % برای ورودی UTF-8
\usepackage{fontspec}
\usepackage{polyglossia}
\usepackage{bidi} % فقط برای کنترل جهت قوی‌تر

\setdefaultlanguage{english}
\setotherlanguage{farsi} % یا persian

\newfontfamily\persianfont[
  Script=Arabic,
  Language=Persian,
  Renderer=HarfBuzz,    % اگر TeX Live قدیمی داری و ارور داد، این گزینه را حذف کن
  Path=./,
  Scale=MatchLowercase
]{Vazirmatn-Light.ttf}

\newcommand{\fa}[1]{{\persianfont\RL{#1}}}

\newcommand{\deepseek}{\textsc{DeepSeek-V3-0324}}
\newcommand{\oss}{\textsc{GPT-OSS-120B}}
\newcommand{\qwen}{\textsc{Qwen3-235B-A22B-Instruct-2507}}
\newcommand{\matina}{\textsc{MatinaRoberta}}

\title{PersianAnonymizer: Evaluating LLM-Labeled Training for Efficient NER-based Anonymization in Persian}

\name{
Mohammad Hossein Shalchian$^{1}$, Mostafa Amiri$^{2}$, Amir Mahdi Sadeghzadeh$^{1}$
}

\address{
$^{1}$Sharif University of Technology\\
\texttt{mo.shalchian@sharif.edu}, \texttt{sadeghzadeh@sharif.edu}\\
$^{2}$University of Tehran\\
\texttt{mostafa.amiri@ut.ac.ir}
}

\abstract{
We target practical anonymization of Persian customer chats by training a compact NER model from LLM-labeled supervision and selecting the best labeler for deployment. We compare three instruction-tuned LLMs---\deepseek, \oss, and \qwen---to produce span annotations under a shared JSON protocol, yielding four corpora (\texttt{OSS\_ZeroShot}, \texttt{Qwen\_ZeroShot}, \texttt{Qwen\_FewShot}, \texttt{DeepSeek\_FewShot}). A \matina-based token-classifier is trained per corpus and evaluated with token-level Precision/Recall/F1 (overall and per-class). We also report \emph{Label Coverage Recall} (LCR), the proportion of gold non-\texttt{O} tokens predicted as non-\texttt{O}, and quantify cross-labeler behavior via a token-level Venn on test annotations. Finally, we contrast test-set annotation latency of the LLMs on H200 nodes with the trained NER’s test-time labeling on a single RTX~3090. Results show that supervision from \texttt{OSS\_ZeroShot} yields the strongest macro-F1 and LCR, while the resulting NER labels an entire 40K-message test set in $\sim$2 minutes on one consumer GPU. This establishes a practical path to high-quality, low-cost anonymization for Persian industrial data.
\\ \newline \Keywords{Named Entity Recognition, Text Anonymization, Persian, Low-Resource, LLM-labeled Supervision, Industrial Chats, Personally Identifiable Information}
}

\begin{document}
\maketitleabstract
\renewcommand{\refname}{ }

\pagestyle{empty}

\section{Introduction}

Organizations increasingly rely on large language models (LLMs) to analyze customer chats and to automate responses, yet these interactions routinely contain personally identifiable information (PII) and, in some domains, protected health information (PHI). Sharing raw messages with third-party or closed-source models raises privacy and compliance concerns (e.g., GDPR), and even aligned language models can memorize and surface snippets from their training data, exacerbating leakage risks \citep{gdpr, carlini2021usenix, shokri2017sp}. A natural mitigation is to \emph{anonymize} inputs before any downstream processing.

\begin{figure}[!t]
  \centering
  \includegraphics[width=1\columnwidth]{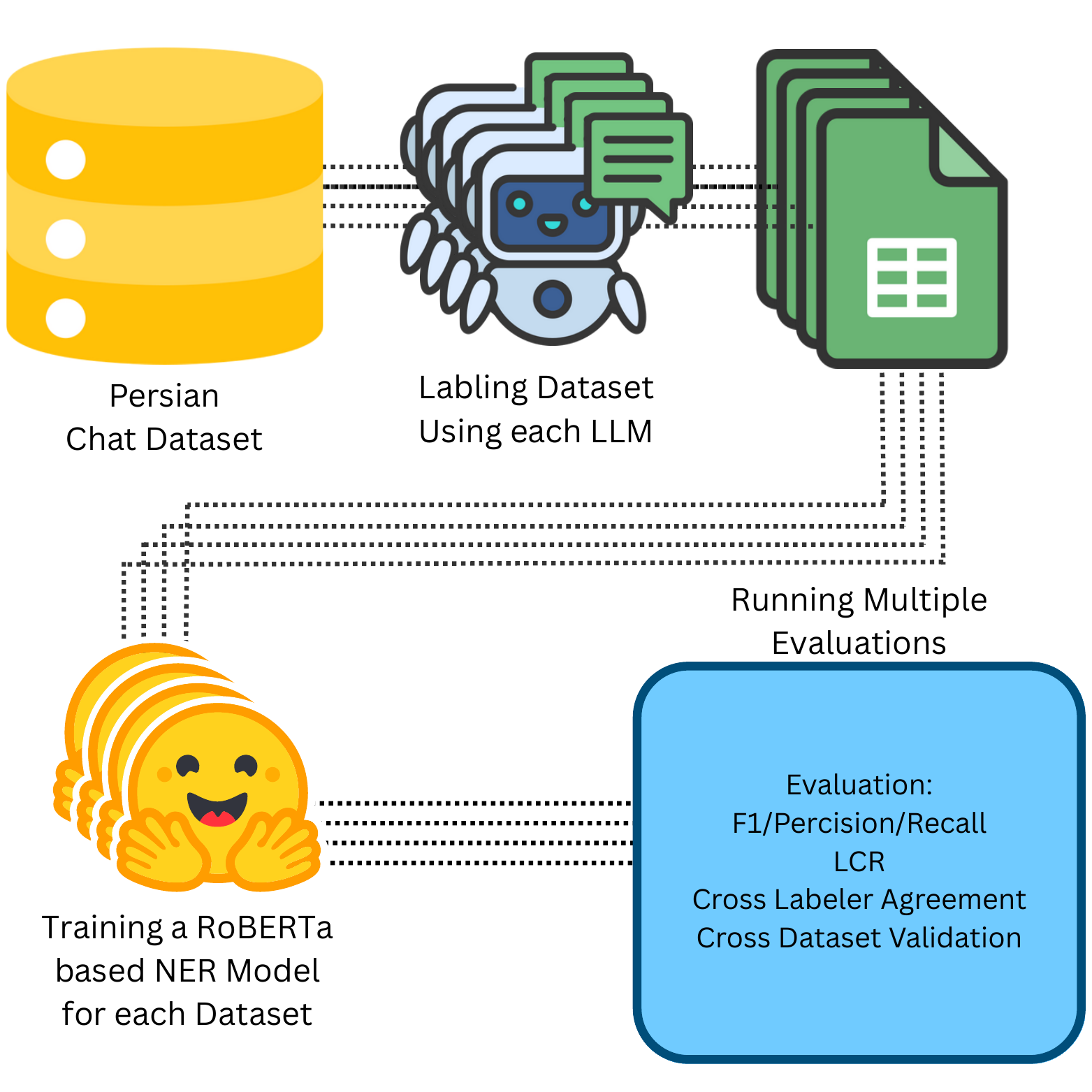}
  \caption{Overall research pipeline: from raw Persian chat data labeling by multiple LLMs to NER training and multi-stage evaluation.}
  \label{fig:pipeline}
\end{figure}

A straightforward path is to ask an LLM to perform anonymization directly; however, production use often encounters a cost–latency–privacy trade-off: high-capacity models demand multi-GPU resources and vendor exposure, while smaller open models may not deliver the needed quality within tight latency budgets. In contrast, a compact \emph{token-classification NER} model, once trained, can anonymize at scale with low, predictable inference cost—provided that we can obtain reliable supervision for the domain.

Recent practice leverages LLMs as \emph{labelers} to bootstrap task models, reducing human annotation overhead \citep{llmaaa_emnlp23}. For Persian, NER resources exist but are comparatively limited and not tailored to industrial chat anonymization. Moreover, to our knowledge there is no systematic comparison of \emph{which} LLM produces supervision that yields the most learnable Persian NER for anonymization scenarios, nor an analysis that balances accuracy with deployment-time efficiency. Recent evidence also documents persistent performance gaps for LLMs in Persian and the need for targeted evaluation, underscoring the importance of our Persian-focused study. \citep{abaskohi-etal-2024-benchmarking}
 Multilingual backbones (e.g., XLM-RoBERTa) \citep{conneau2020xlmr} have advanced transfer to lower-resource languages, but the choice of supervising labeler remains a critical—and underexplored—factor for downstream quality and throughput.

\textbf{In this paper} We compare three instruction-tuned LLMs as automatic labelers for Persian anonymization NER and train a compact NER per labeler. We evaluate token-level precision/recall/F1 (overall and per-class), cross-labeler agreement on test annotations, and end-to-end efficiency. Our results show that labels from one model yield the most learnable supervision, while the resulting NER achieves fast, single-GPU anonymization suitable for industrial deployment.
\textbf{The remainder of this paper is organized as follows:} Section~2 reviews related work on anonymization, LLM-assisted labeling, and Persian NER. Section~3 describes dataset construction and annotation procedures, while Section~4 outlines the NER training setup. Section~5 presents the evaluation results and cross-dataset validation. Finally, Sections~6–8 discuss conclusions, limitations, and ethical considerations.

\section{Related Work}

\subsection{Text Anonymization and De-identification with NER}
Automatic de-identification commonly frames protected attributes, such as PHI and PII, as named entities and applies sequence labeling to mask them prior to downstream processing. Recent work stresses \emph{generalization} challenges across domains and institutions, even with strong transformer baselines \citep{yue2023phicon}. Practical pipelines increasingly combine high-recall NER with conservative post-processing to reduce residual leakage \citep{kocaman-ml4h-2023}. Closer to our goal—substituting costly LLM inference at deployment—``LLMs-in-the-loop'' pipelines train compact task-specific models on LLM-labeled corpora for de-identification across languages \citep{gunay-2024-llms-in-the-loop}.
Cross-domain analyses of pseudonymization indicate that PII tagsets vary substantially across application areas and recommend a core-plus-domain-specific schema, which aligns with our operational label design for Persian chats. \citep{szawerna-etal-2024-pseudonymization}
Our schema emphasizes high-utility PII types (e.g., URL, PHONE, EMAIL) while retaining general entities (PERSON/ORG/LOC) necessary for robust anonymization.
Our study follows this pragmatic line but, to our knowledge, is among the first to quantify (i) \emph{which} LLM yields the most learnable Persian labels for anonymization and (ii) the throughput gains of the resulting NER at inference time.

\subsection{LLMs as NER Labelers (Silver Data)}
LLMs have been used to synthesize labels that subsequently train smaller, task-specific models. Beyond fully synthetic text, a prominent direction is using LLMs as \emph{annotators} over real data, with prompt design, example retrieval, and active acquisition to mitigate noise and cost \citep{llmaaa_emnlp23}.
Complementary strategies such as active learning can further reduce supervision cost for transformer-based NER while retaining accuracy. \citep{vacareanu-etal-2024-active}
To counter label noise inherent to weak or distantly supervised settings, robust training with strong-label-guided lottery mechanisms has shown effectiveness and is conceptually compatible with LLM-labeled pipelines. \citep{ma-etal-2024-enhancing}
For NER specifically, formatting the task to better match generative interfaces and adding self-verification substantially improves label usability, yet pure LLM inference still lags strong supervised baselines and can be resource-intensive \citep{wang-2025-gptner}. Our work adopts this “LLM-as-labeler, NER-for-inference’’ paradigm but focuses on Persian chat data for anonymization, comparing multiple LLM labelers under a unified pipeline and evaluating via downstream learnability and coverage.

\subsection{Persian NER}
Persian NER has a long tail of resources and models. Early corpora such as \textsc{PersoNER} \citep{poostchi2016personer} and \textsc{PEYMA} \citep{fazel-2018-peyma} established baselines, later complemented by BERT-based systems adapted to Persian morphology and domain specifics \citep{mohseni-2019-morphobert,taher-2020-beheshti}. Large Persian PLMs (e.g., ParsBERT) further boosted sequence labeling in standard settings \citep{farahani-2021-parsbert}. However, most prior Persian NER efforts target generic entity extraction; work that \emph{specifically} builds Persian anonymization NER from \emph{LLM-labeled} industrial data and analyzes cross-LLM label quality and inference throughput, remains unaddressed. We address this gap by comparing multiple LLM labelers and selecting the NER trained on the most learnable supervision for practical anonymization at scale.

\section{Dataset Construction}  % (Section 3)
\label{sec:data}

\subsection{Overview}
We construct a large-scale Persian anonymization corpus consisting of \textbf{265{,}000} rows, where each row is a single user message from an organizational chat stream.\footnote{All content is confidential; only anonymized and minimal snippets are shown in figures for illustration. No raw customer data is released.} Messages are short, informal, and often include personally identifiable information (PII) such as names, phone numbers, email addresses, URLs, payment identifiers, and free-form references to dates, addresses, and organizations. Our goal is to obtain span-level named entity annotations suitable for training NER models used in privacy-preserving anonymization.

\subsection{Label Set and Tagging Scheme}
We adopt a BIO tagging scheme over a task-specific PII-oriented label set.
In this scheme, each token is tagged as \texttt{B-} (beginning of an entity), \texttt{I-} (inside an entity), or \texttt{O} (outside any entity). The complete tag vocabulary is:
\begin{itemize}
    \item \textbf{Entity types (without BIO prefixes):} 
    \texttt{COST}, \texttt{CREDIT\_CARD}, \texttt{DATETIME}, \texttt{EMAIL}, \texttt{IBAN}, \texttt{IP\_ADDRESS}, \texttt{LOCATION}, \texttt{NUMBER}, \texttt{ORGANIZATION}, \texttt{PASSWORD}, \texttt{PERSON}, \texttt{PHONENUMBER}, \texttt{URL}, \texttt{USERNAME}.
    \item \textbf{Tags:} for each type we use \texttt{B-} and \texttt{I-} prefixes, plus the outside tag \texttt{O}.
\end{itemize}
This design focuses on PII-bearing or PII-adjacent categories that are common in Persian customer support logs. The label set balances coverage (for operational anonymization) with practical learnability (for downstream NER).

\subsection{LLM-Based Annotation Pipeline}
\label{subsec:llm-pipeline}
We annotated the corpus using three instruction-tuned LLM labelers under an identical prompting and parsing protocol:
\begin{itemize}
    \item \deepseek
    \item \oss
    \item \qwen
\end{itemize}
For each message (one row), we passed a standardized prompt that (i) defines the entity schema and guidelines, (ii) requests JSON-formatted output with explicit entity spans, and (iii) enforces strict formatting via a parsed-API contract. The LLM returns a structure with the raw text and a list of entities, each specified by surface \texttt{phrase} and a canonical \texttt{ner\_type}. A typical response schema is:

\small
\begin{verbatim}
{
  "text": "<original message>",
  "named_entities": [
    {
        "phrase": "...",
        "ner_type": "<PERSON>"
    },
    {
        "phrase": "...",
        "ner_type": "<URL>"
    },
    ...
  ]
}
\end{verbatim}
\normalsize

As an illustrative example, for a Persian sentence referring to two people:\\
\begin{quote}
\fa{"سرکار خانم قربانی - سرکار خانم اسدی از توجه شما متشکرم."}\\[-1pt]
\textit{sarkār khānom Ghorbāni – sarkār khānom Asadi az tavajjoh-e shomā motshakeram.}\\[-1pt]
‘Ms. Ghorbani – Ms. Asadi, thank you for your attention.’
\end{quote}
the LLM may return:
\begin{lstlisting}
{
  "text": "(*@\fa{سرکار خانم قربانی - سرکار خانم اسدی از توجه شما متشکرم.}@*)",
  "named_entities": [
    {"phrase": "(*@\fa{قربانی}@*)", "ner_type": "<PERSON>"},
    {"phrase": "(*@\fa{اسدی}@*)",   "ner_type": "<PERSON>"}
  ]
}
\end{lstlisting}

We used the same prompt template and schema across the three labelers to ensure comparability.

\subsection{Prompting Regimes and Final Datasets}
\label{subsec:prompt-variants}
To examine the effect of instruction conditioning on label quality, we experimented with both \textbf{zero-shot} and \textbf{few-shot} prompting regimes. 
The same schema and output contract described in §\ref{subsec:llm-pipeline} were used for all labelers, but the stability of JSON-formatted outputs varied across models:

\begin{itemize}
    \item \textbf{\oss:} Only the zero-shot configuration produced stable, fully parseable annotations. In the few-shot setting, a large portion of responses violated the JSON schema and could not be recovered, so this variant was discarded.
    \item \textbf{\qwen:} Both zero-shot and few-shot prompting successfully generated valid annotations and were retained for downstream training.
    \item \textbf{\deepseek:} Only the few-shot configuration was usable. Zero-shot outputs frequently failed to comply with the required format and were excluded.
\end{itemize}

Consequently, our study comprises four final LLM-labeled datasets:
\begin{enumerate}
    \item \texttt{OSS\_ZeroShot}
    \item \texttt{Qwen\_ZeroShot}
    \item \texttt{Qwen\_FewShot}
    \item \texttt{DeepSeek\_FewShot}
\end{enumerate}
All datasets were post-processed and normalized into the same token-level BIO format described in §\ref{subsec:alignment}, ensuring consistent label semantics for subsequent NER training and evaluation.

\subsection{Tokenization, Alignment, and Recovery of Missing Spans}
\label{subsec:alignment}
To convert span-level outputs to token-level BIO tags, we (i) tokenize each message, and (ii) align LLM-provided \texttt{phrase} spans to token boundaries using a deterministic regex-based matcher with normalization. Alignment accounts for Persian punctuation, quotes, and common tokenization artifacts (e.g., joined quotes or clitics). When a span crosses multiple tokens, the first token receives \texttt{B-<TYPE>} and the remaining tokens receive \texttt{I-<TYPE>}; non-entity tokens are tagged \texttt{O}.

Occasionally, an entity phrase generated by the LLM did not appear verbatim in the original message—an expected artifact of the probabilistic generation process. In these cases, we applied a fuzzy substring matching strategy: the system searched for sub-sequences in the original text with a similarity score of at least 0.6 (60\%) to the predicted phrase. Entities recovered through this fuzzy matching accounted for fewer than 5\% of all annotations. After recovery, over \textbf{99\%} of LLM-identified entities were successfully located and aligned in text. The remaining $\sim$1\% unmatched cases mostly corresponded to spurious tokens in other scripts (e.g., stray English or Chinese fragments generated by the model), a failure mode that systematically characterized by \citet{marchisio-etal-2024-understanding}.

An example projection for the earlier sentence is shown below (abridged):

\begin{lstlisting}
tokens   = ["(*@\fa{سرکار}@*)","(*@\fa{خانم}@*)","(*@\fa{قربانی}@*)","-",
            "(*@\fa{سرکار}@*)","(*@\fa{خانم}@*)","(*@\fa{اسدی}@*)", ...]
ner_tags = ["B-PERSON","I-PERSON","I-PERSON","O",
             "B-PERSON","I-PERSON","I-PERSON", ...]
\end{lstlisting}

This process yields labeled corpora in \emph{token classification} format, suitable for fine-tuning NER models.

\subsection{Annotation Efficiency and Resource Usage}
\label{subsec:efficiency}

All large-scale annotations were executed on identical hardware nodes equipped with \textbf{8$\times$H200 (140\,GB)} GPUs per node. 
For each model, we launched as many parallel instances as available memory permitted:
\begin{itemize}
    \item \textbf{\deepseek:} one instance (8 GPUs)
    \item \textbf{\qwen:} two instances (each using 4 GPUs)
    \item \textbf{\oss:} four instances (each using 2 GPUs)
\end{itemize}

Each full dataset consisted of \textbf{225{,}000 training} and \textbf{40{,}000 test} messages. 
Table \ref{tab:llm_efficiency} reports the total annotation time for each labeler under the parsed-API setup.

\begin{table}[!ht]
\centering
\begin{tabularx}{\columnwidth}{|l|c|c|>{\centering\arraybackslash}X|}
\hline
\textbf{Labeler} & \textbf{Instances} & \textbf{Train} & \textbf{Test} \\
\hline
DeepSeek & 1 & 92\,min & 22\,min \\
\hline
Qwen & 2 & 36\,min & 6.5\,min \\
\hline
OSS & 4 & \textbf{16\,min} & \textbf{4\,min} \\
\hline
\end{tabularx}
\caption{Annotation times for each LLM labeler on 225K training and 40K test messages (8×H200 140 GB nodes).  
Models used were \textbf{DeepSeek-V3-0324}, \textbf{Qwen3-235B-A22B-Instruct-2507}, and \textbf{GPT-OSS-120B}.  
Parsed-API retries occurred for all labelers but remained under 5\%.}
\label{tab:llm_efficiency}
\end{table}

These measurements provide a consistent baseline for later comparison with the much faster inference of the fine-tuned NER models reported in Sections \ref{sec:train} and \ref{sec:eval}.

\subsection{Cross-Labeler Sample Comparison (Qualitative)}
\label{subsec:qualitative}
Before large-scale training, we qualitatively inspected a shared subset of messages to assess agreement patterns among the three labelers. The following anonymized Persian examples illustrate typical agreement and divergence cases.

\begin{enumerate}
\item 
\begin{quote}
\fa{"سلام لطفاً آی‌پی 10.11.12.13 را به دامنه example.ir متصل کنید."}\\[-1pt]
\textit{salām lotfān IP 10.11.12.13 rā be dāmane example.ir motasel konid.}\\[-1pt]
‘Please connect the IP 10.11.12.13 to the domain example.ir.’
\end{quote}

\textbf{LLM Annotations.} 
All three labelers (\oss, \qwen, \deepseek) correctly identified the personal greeting token as \texttt{PERSON}, the numeric string as \texttt{IP\_ADDRESS}, and the web domain as \texttt{URL}. Boundary consistency was nearly perfect, showing strong convergence on structured identifiers.

\item 
\begin{quote}
\fa{"سرویس اینترنت مبین نت استان کهگیلویه و بویراحمد فعال است."}\\[-1pt]
\textit{service internet Mobin Net ostān Kohgiluye va Boyer-Ahmad fa‘āl ast.}\\[-1pt]
‘The Mobin Net internet service is active in Kohgiluyeh and Boyer-Ahmad province.’
\end{quote}

\textbf{LLM Annotations.}
\oss\ and \qwen\ labeled “\fa{مبین نت}” as \texttt{ORGANIZATION} and “\fa{کهگیلویه و بویراحمد}” as \texttt{LOCATION}, maintaining clear boundaries between organization and place. 
\deepseek, in contrast, extended the \texttt{ORGANIZATION} tag to include tokens like “\fa{استان}”, merging the locative phrase into the organizational span,  which is not an optimal decision as it may obscure certain sentence-level information. Boundary decisions are a well-known fragility in NER—small perturbations near entity edges can induce large drops—so boundary-consistent supervision is crucial for stable generalization. \citep{yang-etal-2024-attack}

\item 
\begin{quote}
\fa{"سلام لطفاً با شماره 09123456789 تماس بگیرید و به آقای رضایی اطلاع دهید."}\\[-1pt]
\textit{salām lotfān bā shomāre 09123456789 tamās begirid va be āghāye Rezā‘i etelā dahid.}\\[-1pt]
‘Please call the number 09123456789 and inform Mr. Rezaei.’
\end{quote}

\textbf{LLM Annotations.}
All models consistently detected the phone number as \texttt{PHONENUMBER} and the person name “\fa{رضایی}” as \texttt{PERSON}. 
\oss\ additionally marked “\fa{آقای}” as part of the person span, while \qwen\ and \deepseek\ left it untagged, showing a subtle divergence in boundary treatment for titles.

\item 
\begin{quote}
\fa{"لطفاً برای گزارش با کدهای 456120 و 456128 اقدام فرمایید."}\\[-1pt]
\textit{lotfān barāye gozāresh bā kod-hāye 456120 va 456128 eghdām farmāyid.}\\[-1pt]
‘Please proceed with the report using codes 456120 and 456128.’
\end{quote}

\textbf{LLM Annotations.}
All three labelers correctly identified both numeric strings as \texttt{NUMBER} entities with precise boundaries and no false positives. This case illustrates near-perfect cross-model agreement on clear numerical tokens.
\end{enumerate}

These examples demonstrate that while all models perform robustly on explicit numeric and technical entities, variation arises primarily in the interpretation of institutional or locative expressions. These qualitative differences informed our downstream evaluation, where model learnability serves as a proxy for label consistency and quality across Persian anonymization corpora.

\section{Training}
\label{sec:train}

We fine-tune a token-classification model based on \matina{} \citep{bourbour2025matina}, 
which itself is derived from XLM-RoBERTa Large \citep{conneau2020xlmr}. 
A linear classification head is (re)initialized to match the BIO tag inventory described in §\ref{sec:data}. 
All four LLM-labeled corpora (\texttt{OSS\_ZeroShot}, \texttt{Qwen\_ZeroShot}, \texttt{Qwen\_FewShot}, \texttt{DeepSeek\_FewShot}) are trained with identical hyperparameters to ensure comparability.

\paragraph{Data splits.}
For each corpus, we use the provided \texttt{train} portion (225K messages) and construct a stratified development split by holding out 10\% of the training data. 
The corresponding \texttt{test} portion (40K messages) is kept untouched for final reporting.

\paragraph{Optimization and schedule.}
We train with AdamW \citep{loshchilov2019adamw}, learning rate $2{\times}10^{-5}$, linear warmup (fixed steps), mixed precision (FP16), and gradient clipping. 
Batch size is kept constant across runs, and training proceeds up to $6$ epochs with early stopping computed on the held-out development split using the \texttt{seqeval} library \citep{seqeval}.
Each run is executed on a single RTX~3090 (24\,GB) and takes about 4 hours.(Table \ref{tab:hparams})

\begin{table}[!ht]
\centering
\caption{Training hyperparameters used identically across the four corpora.}
\begin{tabularx}{\columnwidth}{|l|X|}
\hline
\textbf{Setting} & \textbf{Value} \\
\hline
Backbone & \matina{} \\
\hline
Head & Linear token-classification over BIO tags \\
\hline
Optimizer & AdamW \\
\hline
Learning rate & $2\times 10^{-5}$; linear decay; fixed warmup steps \\
\hline
Batch / Precision & Fixed per run; FP16 \\
\hline
Max epochs & 6 (early stopping on dev macro-F1) \\
\hline
Max sequence length & 256 \\
\hline
Hardware & 1$\times$RTX 3090 (24 GB); about 4h per run \\
\hline
\end{tabularx}
\label{tab:hparams}
\end{table}

\section{Evaluation}
\label{sec:eval}

% -----------------------------------------------------
\begin{table*}[!ht]
\centering
\caption{Macro-averaged token-level results and Label Coverage Recall (LCR, \%).}
\begin{tabular}{|l|c|c|c|c|}
\hline
\textbf{Model (Trained on)} & \textbf{Macro P} & \textbf{Macro R} & \textbf{Macro F1} & \textbf{LCR (\%)} \\
\hline
NER\_OSS (OSS\_ZeroShot)          & \textbf{0.849} & \textbf{0.858} & \textbf{0.851} & \textbf{90.04} \\
\hline
NER\_Qwen (Qwen\_ZeroShot)        & 0.782 & 0.751 & 0.762 & 89.99 \\
\hline
NER\_Qwen (Qwen\_FewShot)         & 0.763 & 0.769 & 0.756 & 89.93 \\
\hline
NER\_DeepSeek (DeepSeek\_FewShot) & 0.768 & 0.720 & 0.733 & 80.04 \\
\hline
\end{tabular}
\label{tab:macro_overview}
\end{table*}

\begin{figure*}[!ht]
  \centering
  \includegraphics[width=\textwidth]{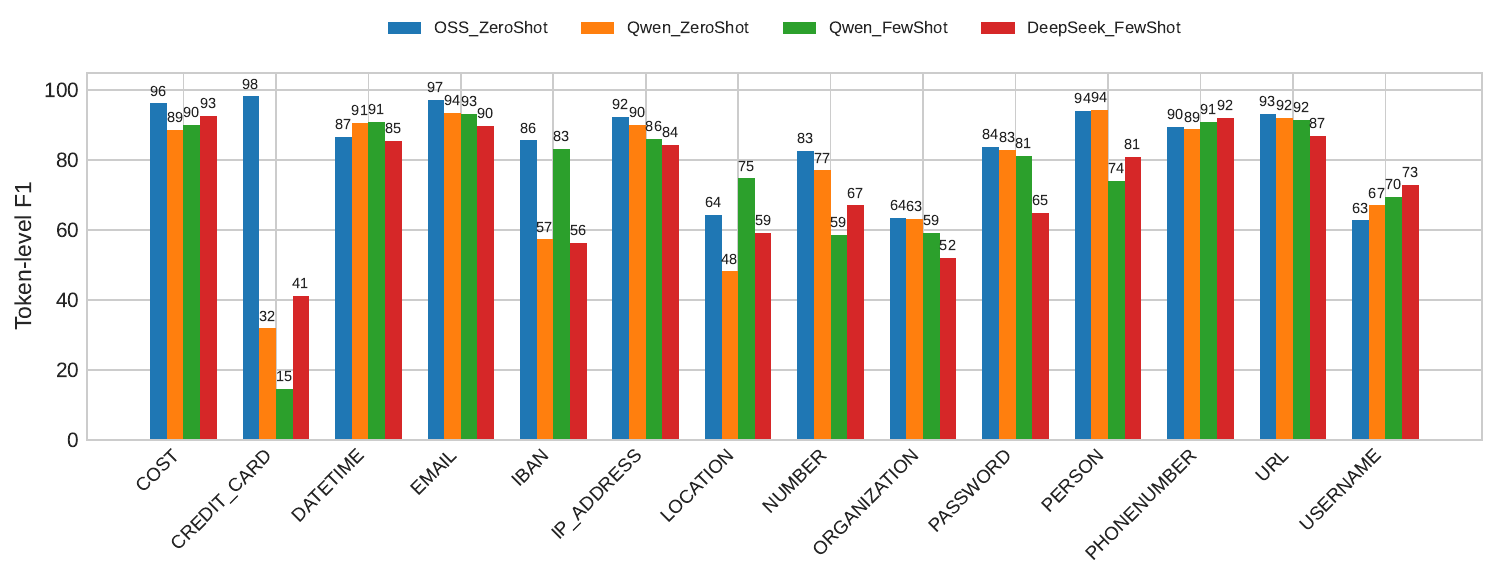}
  \caption{Per-class token-level F1 for the four NER models trained on \texttt{OSS\_ZeroShot}, \texttt{Qwen\_ZeroShot}, \texttt{Qwen\_FewShot}, and \texttt{DeepSeek\_FewShot}. Higher bars indicate better class-wise performance; note that low-support types (e.g., \texttt{CREDIT\_CARD}, \texttt{IBAN}) exhibit higher variance and should be interpreted cautiously.}
  \label{fig:perclass_f1}
\end{figure*}

\begin{figure}[!ht]
\centering
\includegraphics[width=.9\columnwidth]{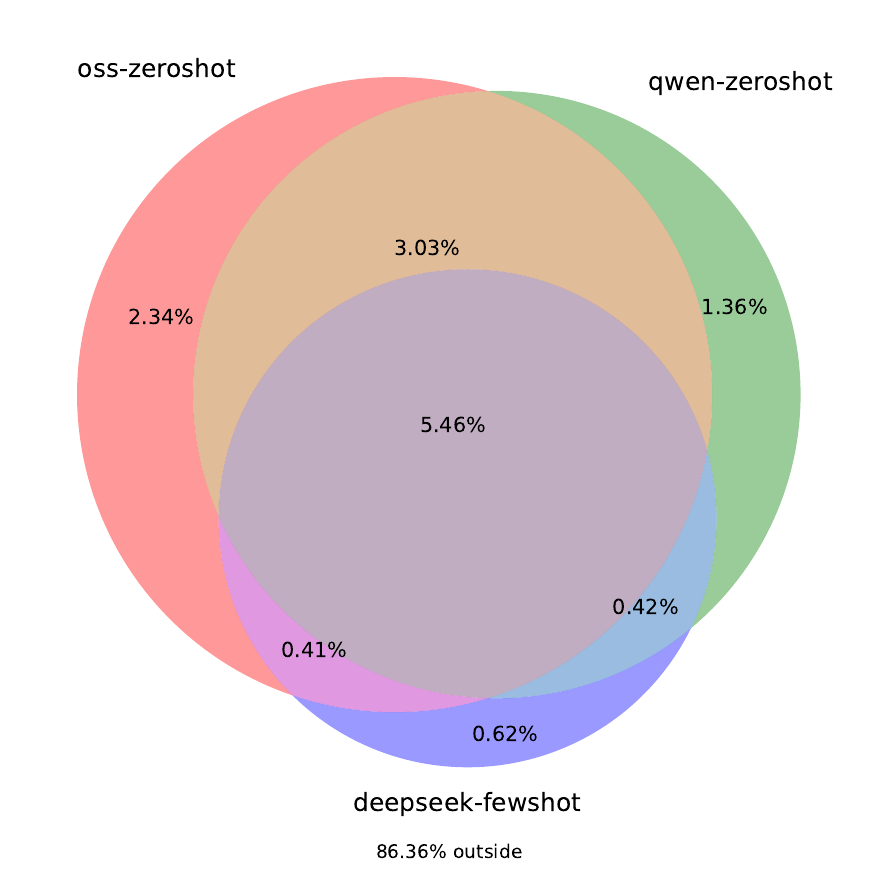}
\caption{Token-level overlap of non-\texttt{O} annotations across the three LLM-labeled test sets: \oss, \qwen~(zero-shot), and \deepseek~(few-shot). Values denote percentage of total tokens.}
\label{fig:venn_labelers}
\end{figure}

% -----------------------------------------------------

\paragraph{Protocol.}
We evaluate the four NER models trained on the LLM-labeled corpora (\texttt{OSS\_ZeroShot}, \texttt{Qwen\_ZeroShot}, \texttt{Qwen\_FewShot}, \texttt{DeepSeek\_FewShot}) using \emph{token-level} counts: for each entity type, gold tokens are those labeled as a non-\texttt{O} tag in the test set, and predictions are computed from the model’s token outputs. We report per-class Precision/Recall/F1 and aggregate macro score. In addition, we introduce \textbf{Label Coverage Recall (LCR)}---the proportion of gold non-\texttt{O} tokens that are predicted as non-\texttt{O} by the model (higher is better). Overall accuracy including \texttt{O} is reported but not emphasized, as it can be misleadingly high in skewed token distributions.
\noindent
Beyond in-domain tests, we also conduct a \textbf{cross-dataset validation} on the \textsc{PEYMA} benchmark \citep{fazel-2018-peyma}, comparing our best NER (trained on \texttt{OSS\_ZeroShot}) against the widely used \textsc{ParsBERT}-NER \citep{farahani-2021-parsbert}. We map overlapping     labels (\texttt{person}, \texttt{organization}, \texttt{location}, \texttt{date}/\texttt{time}$\rightarrow$\texttt{datetime}, \texttt{money}$\rightarrow$\texttt{cost}), ignore BIO prefixes, and evaluate both models on the \emph{shared-label slices} of each test set.

\subsection{Comparison of Trained NER Models}
\label{subsec:ner-comparison}

Table~\ref{tab:macro_overview} summarizes macro-averaged Precision/Recall/F1 and the Label Coverage Recall (LCR) across the four NER models. As seen, the model trained on \texttt{OSS\_ZeroShot} achieves the highest macro-F1 and the strongest LCR, indicating both higher class-balanced quality and broader coverage of entity-bearing tokens.

\noindent

\textbf{Per-class analysis.}
Figure~\ref{fig:perclass_f1} (grouped bars) breaks down token-level F1 by entity type for the four models. Broadly, high-frequency technical identifiers such as \texttt{URL}, \texttt{IP\_ADDRESS}, \texttt{EMAIL}, and \texttt{PHONENUMBER} show consistently strong performance across models, while \texttt{LOCATION} and \texttt{ORGANIZATION} remain comparatively challenging---a known pain point in informal Persian where institutional and locative cues may intertwine.

Some entity types appear very rarely in our test data. For example, \texttt{CREDIT\_CARD} and \texttt{IBAN} each occur only a few dozen times across all datasets (from about 30 examples in OSS to fewer than 200 in Qwen). Because of this small sample size, the reported F1 values for these classes are not statistically reliable and can fluctuate with minor prediction changes. In our target domain—organizational support tickets—such entities are uncommon, so weaker stability on them is not a major concern for the anonymization system as a whole.

Overall, the \texttt{OSS\_ZeroShot}-trained NER not only tops macro-F1 but also attains the best LCR, reflecting better entity-token coverage. The two Qwen variants cluster next, with \texttt{Qwen\_ZeroShot} slightly ahead in macro-F1, and \texttt{DeepSeek\_FewShot} trails due to weaker recall on several mid/high-frequency categories (e.g., \texttt{ORGANIZATION}, \texttt{NUMBER}).

\subsection{Cross-Dataset Validation on \textsc{PEYMA} and Our Test Set}
\label{subsec:cross-peyma}

We align overlapping tags between our schema and \textsc{PEYMA} \citep{fazel-2018-peyma} by mapping \texttt{money}$\rightarrow$\texttt{COST}, \texttt{date}/\texttt{time}$\rightarrow$\texttt{DATETIME}, and keeping \texttt{PERSON}, \texttt{ORGANIZATION}, \texttt{LOCATION}. BIO prefixes are ignored. We then evaluate (i) \textbf{ParsBERT-NER} \citep{farahani-2021-parsbert} and (ii) our \textbf{NER\_OSS} (trained only on \texttt{OSS\_ZeroShot}) on both \emph{PEYMA} and \emph{OSS-labeled} test sets, each restricted to the shared-label slice. For ParsBERT on PEYMA, both \texttt{date} and \texttt{time} tags are merged into a single \texttt{DATETIME} category.

\begin{table*}[!ht]
\centering
\caption{Cross-dataset evaluation on \textsc{PEYMA} (shared labels). Higher is better.}
\begin{tabularx}{\textwidth}{lcccccc}
\hline
\textbf{Label} & \multicolumn{3}{c}{\textbf{NER\_OSS}} & \multicolumn{3}{c}{\textbf{ParsBERT-NER}} \\
 & P & R & F1 & P & R & F1 \\
\hline
COST          & 0.9570 & \textbf{0.9468} & \textbf{0.9519} & \textbf{1.0000} & 0.8370 & 0.9112 \\
DATETIME      & 0.8409 & 0.6743 & 0.7484 & \textbf{0.9570} & \textbf{0.7334} & \textbf{0.8295} \\
LOCATION      & 0.7842 & 0.8530 & 0.8172 & \textbf{0.9482} & \textbf{0.9916} & \textbf{0.9694} \\
ORGANIZATION  & 0.8332 & 0.7066 & 0.7647 & \textbf{0.9573} & \textbf{0.9645} & \textbf{0.9609} \\
PERSON        & 0.8470 & 0.9503 & 0.8957 & \textbf{0.9917} & \textbf{0.9905} & \textbf{0.9911} \\
\hline
\textbf{Macro avg.} & 0.8525 & 0.8262 & 0.8356 & \textbf{0.9708} & \textbf{0.9034} & \textbf{0.9324} \\
\hline
\end{tabularx}
\label{tab:peyma_eval}
\end{table*}

On \textsc{PEYMA} dataset, \textsc{ParsBERT}-NER unsurprisingly leads across most categories and macro averages, due to being trained on multiple datasets including PEYMA. Notably, our \textbf{NER\_OSS} attains stronger \emph{COST} recall and F1, showing competitive transfer on a numeric/structured type.

\begin{table*}[!ht]
\centering
\caption{Cross-dataset evaluation on \textbf{OSS-labeled} test set (shared labels). Higher is better.}
\begin{tabularx}{\textwidth}{lcccccc}
\hline
\textbf{Label} & \multicolumn{3}{c}{\textbf{NER\_OSS}} & \multicolumn{3}{c}{\textbf{ParsBERT-NER}} \\
 & P & R & F1 & P & R & F1 \\
\hline
COST          & \textbf{0.9847} & \textbf{0.9840} & \textbf{0.9844} & 0.9620 & 0.0816 & 0.1504 \\
DATETIME      & \textbf{0.9045} & \textbf{0.9241} & \textbf{0.9142} & 0.6846 & 0.0786 & 0.1411 \\
LOCATION      & \textbf{0.6789} & \textbf{0.6726} & \textbf{0.6757} & 0.3619 & 0.4025 & 0.3811 \\
ORGANIZATION  & \textbf{0.6666} & \textbf{0.6257} & \textbf{0.6455} & 0.5934 & 0.2101 & 0.3103 \\
PERSON        & 0.9298 & \textbf{0.9651} & \textbf{0.9471} & \textbf{0.9352} & 0.5308 & 0.6772 \\
\hline
\textbf{Macro avg.} & \textbf{0.8329} & \textbf{0.8343} & \textbf{0.8334} & 0.7074 & 0.2607 & 0.3320 \\
\hline
\end{tabularx}
\label{tab:oss_eval_vs_parsbert}
\end{table*}

Conversely, on our OSS-labeled test set (restricted to shared labels), \textbf{NER\_OSS} dominates across the board, with particularly large recall gaps on \texttt{DATETIME} and \texttt{ORGANIZATION}. \textsc{ParsBERT}-NER maintains a slight edge in \texttt{PERSON} precision, but overall performance is constrained by low recall—consistent with domain mismatch and the absence of our data in its training mix. Taken together, these two evaluations show expected on-domain advantage for each model, but also support our choice of \textbf{NER\_OSS} as the practical anonymizer for our deployment scenario.

\subsection{Cross-Labeler Agreement on Test Sets (Token-Level Venn)}
\label{subsec:venn}

To examine how consistently the large language models labeled tokens during test-set annotation, we compared the three principal labelers: \textbf{\oss}, \textbf{\qwen} (zero-shot), and \textbf{\deepseek} (few-shot).  
For this analysis, we took the respective \texttt{test} sets produced by each LLM and computed, at the token level, whether each token was assigned a non-\texttt{O} label by each labeler.  
In other words, we treat every token predicted as part of any entity (regardless of its specific class) as a ``labeled'' token, and every \texttt{O} as ``unlabeled.''  
The resulting binary indicators across the three datasets yield a 3-set overlap pattern visualized as a Venn diagram in Figure~\ref{fig:venn_labelers}.  

\noindent
Table~\ref{tab:venn_stats} summarizes the distribution of token-labeling coverage and overlap percentages.

\begin{table}[!ht]
\begin{center}
\begin{tabularx}{\columnwidth}{|l|c|}
\hline
\textbf{Subset (labelers)} & \textbf{Percent (\%)} \\
\hline
Only GPT-OSS (Zero-Shot) & 2.34 \\
Only Qwen3 (Zero-Shot) & 1.36 \\
Only DeepSeek (Few-Shot) & 0.62 \\
GPT-OSS $\cap$ Qwen3 & 3.03 \\
GPT-OSS $\cap$ DeepSeek & 0.41 \\
Qwen3 $\cap$ DeepSeek & 0.42 \\
GPT-OSS $\cap$ Qwen3 $\cap$ DeepSeek & 5.46 \\
None (all \texttt{O}) & 86.36 \\
\hline
\end{tabularx}
\caption{Token-level overlap of non-\texttt{O} labels across the three LLM-labeled test sets. Percentages are computed over all tokens in the union of test corpora.}
\label{tab:venn_stats}
\end{center}
\end{table}

\noindent
Roughly $13.6\%$ of all tokens received a non-\texttt{O} label from at least one LLM.  
The highest overlap region (\textbf{5.46\%}) corresponds to tokens that all three labelers agreed were part of some entity, while exclusive regions (e.g., only-\oss{} 2.34\%) reveal labeler-specific biases.  
Overall, \oss{} labeled the largest fraction of tokens, and its intersections with the other two models dominate the shared areas—consistent with its broader coverage observed in Section~\ref{subsec:ner-comparison}.  
In contrast, \deepseek{} contributed relatively few unique labeled tokens (0.62\%), confirming that it was the most conservative labeler among the three.  
This overlap pattern highlights that while the models converge on many clear entity tokens, a considerable proportion of potential entities remain model-specific, underscoring heterogeneity in LLM-based annotation behavior.

\subsection{Practical Throughput: NER vs.\ LLM Labeling}
\label{subsec:efficiency_eval}

While the large language models required substantial multi-GPU resources to annotate the test sets (§\ref{subsec:efficiency}), the fine-tuned NERs perform inference extremely fast.  
Each trained model labels the entire 40K-message test set in roughly \textbf{2 minutes} on a single RTX~3090 (24 GB), compared to tens of minutes on multi-node H200 setups for the original LLMs.  
This demonstrates that high-quality anonymization can be achieved with a lightweight model at a fraction of the computational cost—offering a practical balance between \emph{label fidelity} and \emph{deployment efficiency}.

\section{Conclusion}

Our findings indicate that supervision from \texttt{OSS\_ZeroShot} consistently yields the most learnable Persian anonymization NER, balancing class-wise quality (macro-F1) with broad entity-token coverage (LCR). The token-level Venn further shows sizable but incomplete cross-labeler agreement, suggesting that labeler heterogeneity remains a meaningful source of variance. Practically, a compact NER trained on this supervision labels a 40K-message test set in ~2 minutes on a single RTX 3090—making privacy-preserving preprocessing feasible under tight latency and cost budgets.
For deployment, we recommend prioritizing OSS-labeled supervision and complementing the NER with lightweight heuristics for challenging types (e.g., ORG/LOC boundary cues) where recall matters most. Future work includes human audits on sampled segments, agreement-aware label consolidation, and evaluating transfer to adjacent Persian domains.

\section{Ethics Statement}

All chat data were processed under explicit organizational authorization for research on privacy-preserving anonymization. The data were used solely for non-commercial, internal research; no raw customer messages are released. All examples shown in the paper are anonymized: identifying strings are replaced with semantically similar surrogates, and any potentially sensitive details are removed or perturbed.

Because of confidentiality and data-protection commitments, the underlying chat datasets cannot be shared. We may release the trained NER model checkpoint(s) since the model is task-specific (token classification) and does not generate text; this reduces the risk of reproducing original messages. We further ensured that no raw texts are embedded in code, logs, or artifacts released with the paper.

\section{Limitations}

Our study has several limitations. 
(i) \textbf{Recall is not perfect.} Although overall LCR is high, recall is lower on certain categories (e.g., \texttt{ORGANIZATION}, \texttt{LOCATION}), and boundary errors persist in informal, code-mixed contexts. 
(ii) \textbf{Sparse classes.} Some entity types (e.g., \texttt{CREDIT\_CARD}, \texttt{IBAN}) have very few instances; their per-class F1 is variance-prone and should be interpreted with caution. This reflects our target domain (enterprise support tickets) rather than an inherent weakness of the approach. 
(iii) \textbf{Label noise and evaluation scope.} Supervision is produced by LLMs; while downstream learnability is a useful proxy, we lack large-scale human adjudication. Our token-level Venn aggregates all non-\texttt{O} labels and ignores type identity, so it measures coverage overlap rather than class agreement. 
(iv) \textbf{Domain specificity.} Data come from a single organizational setting; performance may degrade under distribution shift (different industries, styles, or policies).

% ----------------- References -----------------

\section{Bibliographical References}\label{sec:reference}

\bibliographystyle{lrec2026-natbib}
\bibliography{refs}

\begin{thebibliography}{8}
\expandafter\ifx\csname natexlab\endcsname\relax\def\natexlab#1{#1}\fi

\bibitem[{AI(2025)}]{deepseek2025}
DeepSeek AI. 2025.
\newblock \href {https://huggingface.co/deepseek-ai/DeepSeek-V3-0324}
  {\emph{DeepSeek-V3-0324}}.
\newblock Used for dataset annotation (few-shot).

\bibitem[{Cohere(2025)}]{qwen2025}
Cohere. 2025.
\newblock \href {https://huggingface.co/Qwen/Qwen3-235B-A22B-Instruct-2507}
  {\emph{Qwen3-235B-A22B-Instruct-2507}}.
\newblock Used for dataset annotation (zero/few-shot).

\bibitem[{et~al.(2020)}]{xlmr2020}
Alexis Conneau et al. 2020.
\newblock \href {https://huggingface.co/xlm-roberta-large} {\emph{XLM-RoBERTa:
  A Strong Baseline for Cross-lingual Understanding}}.
\newblock Base multilingual transformer used for Matina.

\bibitem[{Farahani et~al.(2021)Farahani, Ghassemi, Farahani
  et~al.}]{farahani-2021-parsbert}
Farahani, Mohammad Mahdi and Ghassemi, Naser and Farahani, Mohammad Ali and
  others. 2021.
\newblock \href
  {https://huggingface.co/HooshvareLab/bert-base-parsbert-ner-uncased}
  {\emph{ParsBERT: Transformer-based Model for Persian Language
  Understanding}}.
\newblock Used model for evaluation.

\bibitem[{Hosseinbeigi et~al.(2025)Hosseinbeigi, Asghari, Kashani, Shalchian,
  and Abbasi}]{matina2025}
Sara Bourbour Hosseinbeigi and Sina Asghari and Mohammad Ali Seif Kashani and
  Mohammad Hossein Shalchian and Mohammad Amin Abbasi. 2025.
\newblock \href {https://huggingface.co/MatinaAI} {\emph{MatinaRoberta: A
  Persian Language Model Derived from XLM-RoBERTa}}.
\newblock Pretrained model used as NER backbone.

\bibitem[{OpenAI(2024)}]{oss2024}
OpenAI. 2024.
\newblock \href {https://huggingface.co/openai/gpt-oss-120b}
  {\emph{GPT-OSS-120B}}.
\newblock Used for dataset annotation (zero-shot).

\bibitem[{Organization(2025)}]{persianchat2025}
Anonymous Organization. 2025.
\newblock \emph{Persian Industrial Chat Corpus}.
\newblock Confidential industrial dataset; non-public; used under
  authorization.

\bibitem[{Shahshahani et~al.(2018)Shahshahani, Sameti, Shamsfard
  et~al.}]{fazel-2018-peyma}
Shahshahani, Fatemeh and Sameti, Hossein and Shamsfard, Mehrnoush and others.
  2018.
\newblock \href {https://huggingface.co/datasets/ParsiAI/PEYMA} {\emph{PEYMA: A
  Tagged Corpus for Persian Named Entities}}.
\newblock Used dataset for evaluation.

\end{thebibliography}

\nocitelanguageresource{matina2025,xlmr2020,oss2024,qwen2025,deepseek2025,persianchat2025, fazel-2018-peyma, farahani-2021-parsbert}
\section{Language Resource References}
\label{lr:ref}
\bibliographystylelanguageresource{lrec2026-natbib}
\bibliographylanguageresource{ref2}

\end{document}